\documentclass[10pt,twocolumn,letterpaper]{article}

 \usepackage{cvpr}              % To produce the CAMERA-READY version
\usepackage[accsupp]{axessibility}

\usepackage{natbib}

\usepackage{amsfonts}
\usepackage{graphicx}
\usepackage{array}
\usepackage{booktabs} % For formal tables
\usepackage{bm}
\usepackage{float}
\usepackage{times,amsmath,epsfig}
\usepackage{graphicx}%插入图片
\usepackage{amssymb}%数学符号
\usepackage{amsthm}%数学定理
\usepackage[dvipsnames]{xcolor}
\usepackage{algorithm}
\usepackage{algorithmic}
\usepackage{multirow} %多行合并
\usepackage{array} %\调用公式宏包的命令应放在调用定理宏包命令之前，也能控制表格
\usepackage{booktabs} %调整表格线与上下内容的间隔
\usepackage{longtable}%调用跨页表格
\usepackage{bm}%数学字体加粗
\usepackage{verbatim} %多行注释
\usepackage{float}

\usepackage{color}
\usepackage{verbatim}
\usepackage{tcolorbox}
\newcolumntype{H}{>{\setbox0=\hbox\bgroup}c<{\egroup}@{}}
\usepackage{colortbl}
\usepackage{pifont}
\newcommand{\xmark}{ \ding{55}}%
\usepackage{makecell}

\definecolor{lightgray}{rgb}{0.9, 0.9, 0.9}
\usepackage{arydshln}
\usepackage[dvipsnames]{xcolor}

\setlist[itemize]{itemsep=0pt, parsep=0pt, partopsep=0pt, topsep=0pt}

\usepackage{marvosym}
\definecolor{cvprblue}{rgb}{0.21,0.49,0.74}
\usepackage[pagebackref,breaklinks,colorlinks,citecolor=cvprblue]{hyperref}
\def\paperID{5551} % *** Enter the Paper ID here
\def\confName{CVPR}
\def\confYear{2025}

\title{Towards All-in-One Medical Image Re-Identification}
\vspace{-4mm}
\author{\small   Yuan Tian$^{1}$
	~~
	Kaiyuan Ji$^{2}$
	~~
	Rongzhao Zhang$^{1}$
	~~
		Yankai Jiang$^{1}$
	~~
		Chunyi Li$^{3}$\\
	\small	Xiaosong Wang$^{1}$
	~~
		Guangtao Zhai$^{3}$
	\\
	\small $^{1}$Shanghai AI Laboratory\\
	\small $^{2}$School of Communication and Electronic Engineering, East China Normal University\\
	\small $^{3}$Institute of Image Communication and Network Engineering, Shanghai Jiao Tong Unversity\\
	{\tt\small tianyuan168326@outlook.com}\vspace{-0.7em}		
}

\begin{document}

\maketitle
\begin{abstract}
Medical image re-identification (MedReID) is under-explored so far, despite its critical applications in personalized healthcare and privacy protection.
In this paper, we introduce a thorough benchmark and a unified model for this problem.
First, to handle various medical modalities, we propose a novel Continuous Modality-based Parameter Adapter (ComPA). ComPA condenses medical content into a continuous modality representation and dynamically adjusts the modality-agnostic model with modality-specific parameters at runtime. This allows a single model to adaptively learn and process diverse modality data.
Furthermore, we integrate medical priors into our model by aligning it with a bag of pre-trained medical foundation models, in terms of the differential features.
Compared to single-image feature, modeling the inter-image difference better fits the re-identification problem, which involves discriminating multiple images.
We evaluate the proposed model against 25 foundation models and 8 large multi-modal language models across 11 image datasets, demonstrating consistently superior performance.
Additionally, we deploy the proposed MedReID technique to two real-world applications, i.e., history-augmented personalized diagnosis and medical privacy protection.
Codes and model is available at \href{https://github.com/tianyuan168326/All-in-One-MedReID-Pytorch}{https://github.com/tianyuan168326/All-in-One-MedReID-Pytorch}.

\end{abstract}    
\vspace{-2mm}
\section{Introduction}
\label{sec:intro}

Medical images~\cite{webb2022introduction}, such as X-ray images and Computed Tomography (CT) scans, are essential for diagnosing and monitoring various health conditions.
Up to 2020, images have accounted for about 90\% of all medical data~\cite{zhou2021review}.

Despite the large-scale data advanced the computer-aided diagnosis tasks~\cite{van2011computer,cai2020review}, its privacy concern~\cite{price2019privacy} is also serious.
It is urgent to (1) efficiently manage patient historical images for personalized healthcare~\cite{lambin2017radiomics, aerts2016potential, panayides2020ai} and (2) effectively protect privacy before images are shared~\cite{giakoumaki2006secure, kim2021privacy, heinrichimplicit}.
We argue that both sides call for the medical image re-identification (MedReID) technique.

\begin{figure}[!tbp]
\centering
\tabcolsep=0.1mm
\begin{tabular}{cc}
	\includegraphics[width=0.99 \linewidth]{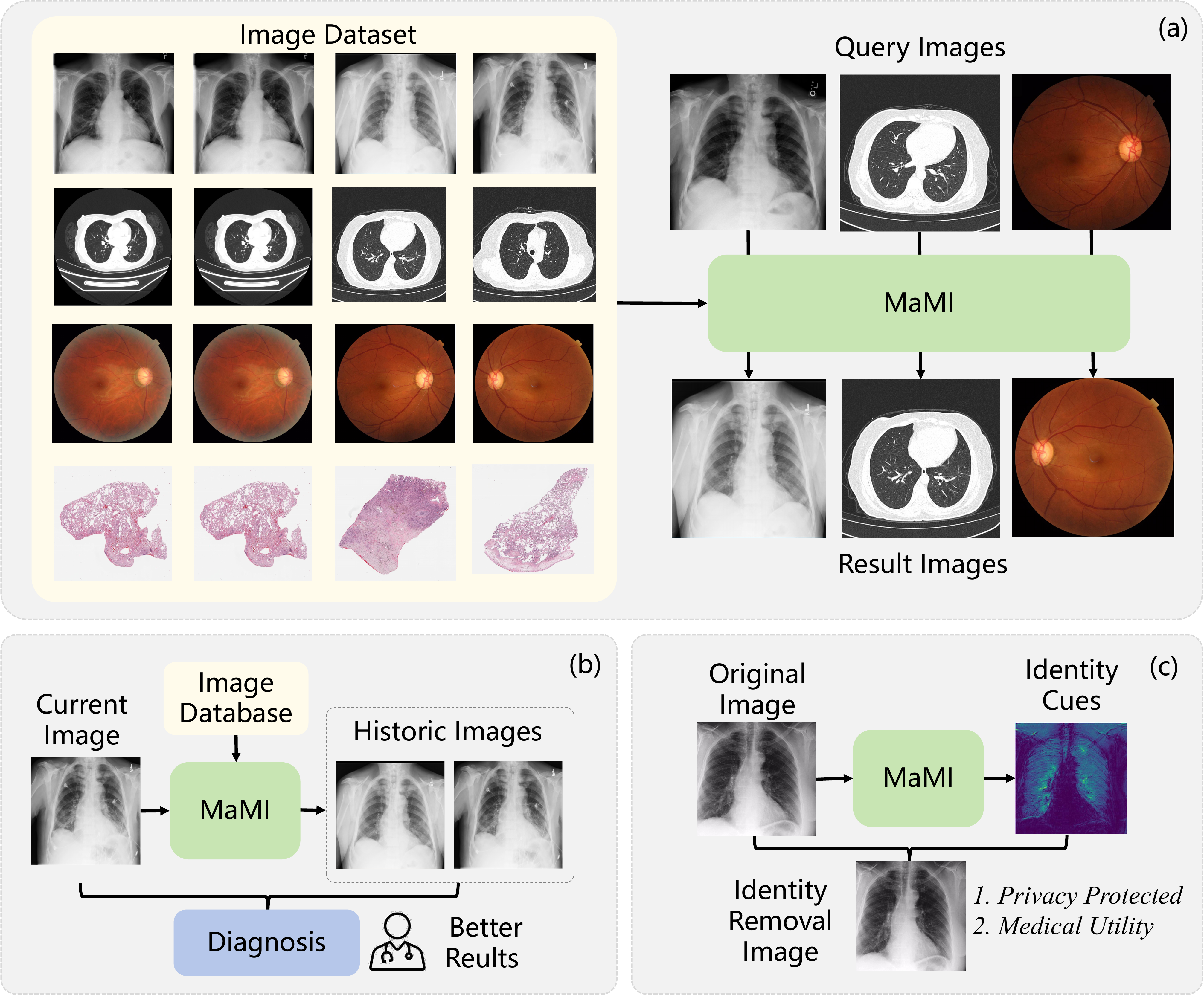}
\end{tabular}
\vspace{-3mm}
\caption{
(a) We propose MaMI, an all-in-one modality-adaptive ReID model for medical images.
(b) MaMI enhances personalized healthcare by integrating historical medical data.
(c) MaMI detects identity cues and removes them from the original images, protecting privacy while maintaining medical utility.
}
\vspace{-4mm}
\label{fig:task_description}
\end{figure}

As for historical image management, traditional methods~\cite{meyer1994picture,teng2010medical} manually pre-link images to patient metadata (e.g., name, medical record numbers), and retrieve images by querying the system with the metadata.
However, the links are not always complete and accurate, especially when data are stored in different Picture Archiving and Communication System (PACS) platforms. 
This requires the MedReID technique to retrieve personal images from poorly organized data, providing accurate historical evidence for disease diagnosis~\cite{jin2021predicting}.

As for medical image privacy protection, current methods only remove explicit information, such as the patient name~\cite{johnson2019mimic}.
However, some works~\cite{packhauser2022deep,ganz2025re} have found that the identifiable visual information within the images can also breach privacy.
A robust MedReID model can detect the identity-related regions of the image. By post-processing these regions, the images become unidentifiable, thereby enhancing their safety before data sharing.

Despite the importance of the MedReID problem, there are only few works investigating this.
Fukuta~\etal~\cite{fukuta2008personal} and Singh~\etal~\cite{singh2016unique} exploit the low-level features for identifying fundus images.
Packhäuser~\etal~\cite{packhauser2022deep} leverages neural networks to identify chest X-ray images.
However, all these approaches are designed for one specific modality.
They can not enjoy the mutual enhancement from various-modality data sources.
Moreover, these models are with less medical priors, which limits their generalization.

In this paper, we introduce a unified MedReID model, termed Modality-adaptive Medical Identifier (MaMI).
To handle heterogeneous data from various modalities, MaMI introduces a Continuous Modality-based Parameter Adapter (ComPA). ComPA adapts a modality-agnostic model to modality-specific models at runtime. Given an input image, ComPA generates a continuous modality context, which dynamically produces modality-specific parameters. These parameters are then used to adjust the modality-agnostic model, enabling accurate re-identification of diverse medical modalities with a single model.

Furthermore, we integrate medical priors into our model by aligning it with pre-trained medical foundation models (MFMs), in terms of the inter-image key feature differences.
The key features are obtained by attending to the local features using a group of learnable modality-specific query tokens.
Compared to the single-image feature, the inter-image differences are more consistent with the ReID, which targets discriminating the identity relation of multiple images.

We compare our model, MaMI, against 25 foundation models and 8 large multi-modal language models across 11 medical image datasets, encompassing a wide range of modalities and body organs, establishing a thorough benchmark for the MedReID problem. Our model consistently outperforms the others.
Additionally, we deploy our approach in real-world applications.
First, historical data-augmented diagnosis, i.e., MaMI retrieves personalized historical patient data from unorganized datasets, significantly enhancing the accuracy of current medical examinations.
Second, privacy protection, i.e., MaMI detects subtle visual cues that reveal patient identity and removes them from images before data sharing, ensuring privacy while preserving medical utility.
Our contributions are:
\begin{itemize}
	\setlength{\itemsep}{0pt}
	\setlength{\parsep}{0pt}
	\setlength{\parskip}{0pt}
	
\item
We propose the first all-in-one medical re-identification model, termed MaMI, capable of re-identifying medical images of various modalities using a single model.
We build a thorough and fair benchmark for this novel problem.
	
\item We propose a novel Continuous modality-based Parameter Adapter, which dynamically produces modality-specific parameters, and enables the model to adaptively re-identify different modalities.

\item Our model inherits the medical priors from medical foundation models, while adapting them to the ReID problem by inter-image difference modeling.

\item We showcase that MaMI can benefit real-world medical applications, e.g., history-augmented healthcare and medical privacy protection.

\end{itemize}

% \vspace{-2mm}
\section{Related Work}
% \vspace{-1mm}
\textbf{Medical Image Re-Identification (MedReID).}
Numerous medical models focus on automatically diagnosing medical images~\cite{wei2024focal, cai2020review} or retrieving the images by disease features~\cite{fang2021deep, jush2024medical}.
There are few works focusing on the MedReID problem. Heinrich \etal~\cite{heinrich2024automatic} utilized low-level image descriptors such as Sobel~\cite{zhao2008sobel} to detect patient identity from head CT images. Packhäuser \etal~\cite{packhauser2022deep} and Ganz \etal~\cite{ganz2025re} re-identify patients from chest X-ray and histopathology images, respectively. However, all these approaches are limited to a single modality and cannot benefit from large-scale data of various modalities.

%To address these gaps, we propose the first all-in-one model capable of re-identifying various modalities. Moreover, our approach explicitly incorporates medical priors, improving the model's robustness and generalizability.

\textbf{Object Re-Identification.}
Most approaches~\cite{ye2024transformer} focus on identifying faces~\cite{zhao2003face,deng2019arcface,parkhi2015deep,tian2024medical}, persons~\cite{hermans2017defense,luo2019bag,he2021transreid,yan2023clip,fu2021unsupervised,chen2023beyond,wang2019spatial,luo2021self,yin2020fine,kim2023partmix,hu2024personmae,he2024instruct}, animals~\cite{rao2021counterfactual,jiao2024toward}, and vehicles~\cite{zheng2020vehiclenet,khan2019survey,zhou2020fine}.
However, there are few methods dedicated to medical images.

\textbf{Medical Foundation Models.}
Early, there are amounts of dedicated models for independent tasks, such as video recognition~\cite{feichtenhofer2019slowfast,tian2024coding,tian2022ean,tian2024free,chen2024gaia,tian2024smcplus,tian2023non,tian2020self}, low-level image processing~\cite{zhang2018residual,zhao2024wavelet,zhao2024cycle,zhao2024spectral,xie2024addsr,yi2021attention,tian2021self,tian2023clsa}, and medical image analysis~\cite{cao2022swin,tian2019video,chen2024cross} tasks.
Later, foundation models~\cite{kirillov2023segment,oquab2023dinov2} are becoming more and more popular, due to their strong generalization capability and strong performance.
Recently, numerous medical foundation models, such as X-ray models~\cite{wang2022multi,yao2024eva,wan2024med}, fundus image models~\cite{zhou2023foundation}\cite{shi2024eyeclip}, and CT models~\cite{hamamci2024foundation_ctclip,wu2023towards}, have been continuously proposed.
We are the first to adapt their medical priors to the MedReID problem.

\textbf{Medical Image Domain Adaptation.}
Medical image domain adaptation addresses domain shifts in imaging data, improving model generalization across different clinical settings~\cite{zeng2021semantic,guan2021domain,zhang2020collaborative}.
However, these methods mainly focus on diagnosis tasks, how to devise a highly generalizable medical ReID model is left blank.

\begin{figure*}[!t]
	%	\vspace{-3mm}
	\centering
	\includegraphics[width=17.5cm]{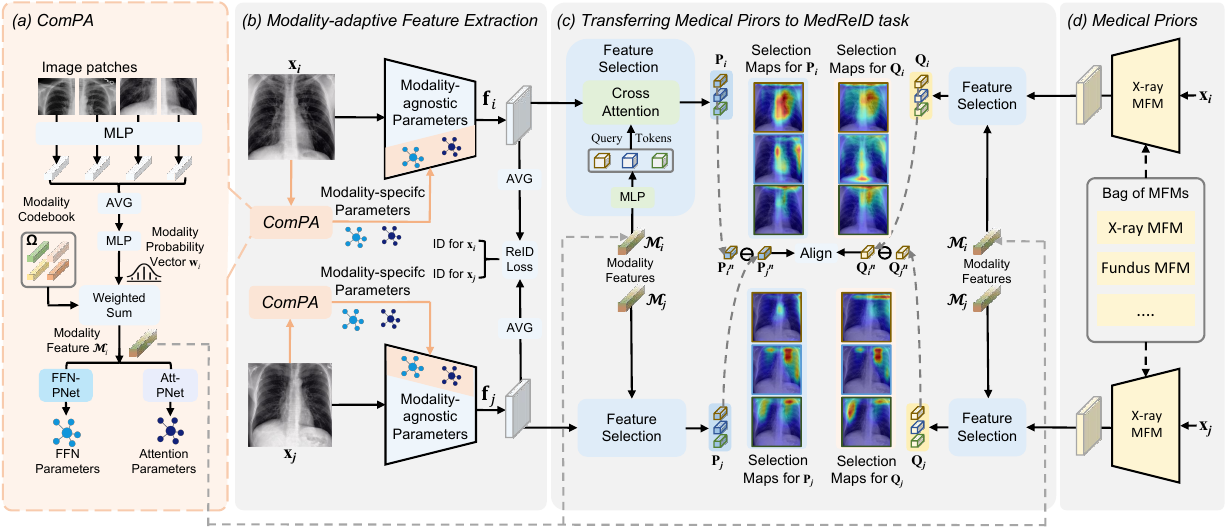}
	\vspace{-5mm}
	\caption {
	   Overview of the proposed all-in-one MedReID framework, namely Modality-adaptive Medical Identifier (MaMI). (a) We introduce a Continuous Modality-based Parameter Adapter (ComPA) to dynamically adjust a modality-agnostic model into an input modality-specific model at runtime. (b) The adjusted model extracts the identity-related visual features from the input medical images.
       (c) During the optimization, we also transfer the rich medical priors from the (d) medical foundation models (MFMs) to the MedReID task, by aligning the inter-image key differences.
       We illustrate with X-ray images, though our method also supports other modalities.}
	\vspace{-5mm}
	\label{fig:framework}
\end{figure*}

% \vspace{-2mm}
\section{Approach}
% \vspace{-1mm}

\subsection{Overview}
As outlined in Figure~\ref{fig:framework}, we introduce two key ideas to enable a single model to identify various-modality medical images, in an all-in-one manner.
First, we achieve modality-adaptive feature extraction, by upgrading a modality-agnostic model to a modality-specific model at runtime.
Second, we optimize the model to focus more on medically-relevant regions, by transferring the medical priors within medical foundation models to the MedReID task.

\subsection{Modality-Adaptive Feature Extraction}
We leverage a typical Transformer network, ViT-Base~\cite{dosovitskiy2020image}, as the backbone for feature extraction. ViT consists of several attention blocks and feed-forward networks (FFNs). During runtime, we dynamically adjust the network to cater to the current input image.

\textbf{Motivation.}
We try to fine-tune a pre-trained ViT model, namely, CLIP~\cite{radford2021learning}, towards the MedReID task with two strategies,  (1) Single-modality, which separately fine-tunes a specialized model for each modality, and (2) Multiple-modality, which combines the data of all modalities and fine-tunes a unified model.
The results are shown in Table~\ref{fig:motivation_mom_lora}.
Compared to single-modality, the multiple-modality strategy shows improvement in eye fundus modality (76.88\% $\rightarrow$ 82.48\%), while demonstrating a decrease in X-ray modality (94.21\% $\rightarrow$ 92.30\%).
This indicates that using combined data to learn a unified model benefits some modalities due to more training data, while also limiting the upper bound of some other modalities.
We argue that the reason is that, naively putting multiple modalities into a single model, mostly learns the modality-agnostic knowledge, neglecting the modality-specific knowledge.

\begin{table}[!thbp]
%		\vspace{-1mm}
	\centering
		\renewcommand{\arraystretch}{0.3}
	\begin{tabular}{c|c|c}
		\hline
		Method & X-ray (\%) & Fundus (\%) \\
		\hline
		CLIP baseline & 33.10 & 41.14 \\
		Single-modality & 94.21 & 76.88 \\
		Multiple-modality & 92.30 & 82.48 \\
        \hline
  	\rowcolor{gray!25}
		Continuous-modality (Ours) & 96.89 & 85.71 \\
		\hline
	\end{tabular}
		\vspace{-2mm}
 \caption{Comparison of different modality handling strategies.
 We adopt the MIMIC-X~\cite{johnson2019mimic} and Mess2~\cite{decenciere2014feedback} datasets to evaluate the performances on X-ray and eye fundus images.
 }
	\label{fig:motivation_mom_lora}
		\vspace{-5mm}
\end{table}

\textbf{Continuous modality-based Parameter Adapter (ComPA).}
To address the above challenge, we propose the ComPA to amend the modality-agnostic model with input-modality-specific model parameters, as shown in Figure~\ref{fig:framework} (b).
This effectively decouples the learning of modality-agnostic and modality-specific knowledge.

Rather than employing categorical modality labels, such as 0/1/2 for X-ray/Fundus/CT, ComPA introduces a novel continuous modality representation to handle the modality specificity, as shown in Figure~\ref{fig:framework} (a).
Specifically, given an input image \( \mathbf{x}_i \in \mathbb{R}^{3 \times H \times W} \), where $H$ and $W$ denote its spatial scales, we convert each \( 16 \times 16 \) patch into local modality contexts by a Multilayer Perceptron (MLP), which are averaged to obtain the global modality context.

To improve generalization for images outside the training domain, instead of directly employing the above un-constrained modality context, we constrain the underlying modality representations to be derived from a set of basis centers. Specifically, another MLP transforms the global context into a modality probability vector \( \mathbf{w} \in \mathbb{R}^{L} \), where \( L = 32 \) denotes the number of all pseudo modalities. Note that this number significantly exceeds the typical number of medical modalities, such as CT and X-ray, due to the diverse imaging styles within a single modality class. For example, variations in X-ray machines and settings can result in numerous imaging styles \cite{bushberg2011essential}. \( \mathbf{w}_i \) is then used to compute a weighted sum of learnable modality bases \( \mathbf{\Omega} \in \mathbb{R}^{L \times 768} \), producing the ultimate continuous modality feature $ \mathcal{M}_i = \mathbf{w}_i\mathbf{\Omega} \in \mathbb{R}^{768} $.
$\mathbf{\Omega}$ is randomly initialized and learned with other components in an end-to-end manner.

Given $\mathcal{M}_i$, two MLPs, named Att-PNet and FFN-PNet, generate parameters for the attention and FFN layers of the ViT model, respectively.
Nevertheless, directly predicting these parameters would require an infeasibly large number of parameters.
For instance, for a ViT-Base model with 86M parameters,  given the dimension of $\mathcal{M}_i$ is 768, the last layer of the above two PNets would include \(86 \times 768\)M \(\approx 66\)G parameters, which is intractable.
To mitigate this issue, we predict low-rank parameters~\cite{hu2021lora}, instead of the full parameters. Meanwhile, we implement the last linear layer of two PNets in a group-wise manner~\cite{xie2017aggregated}, for further reducing the parameter number and computational cost.

The above-generated parameters are merged into the modality-agnostic network in a layer-wise manner. Following LORA~\cite{hu2021lora}, we expand the generated low-rank parameters to match the shape of the ViT layers and add them to the corresponding layer parameters.

Our approach
shares similarities with recent Mixture-of-Expert (MOE)-LORA paradigms~\cite{wumixture,yang2024multi}, which dynamically weights a series of LORA modules. However, there are two fundamental differences. \textit{Goal Difference}: We aim to perceive the input medical image modality by operating on low-level patch features, whereas MOE-LORA methods utilize high-level semantic features to select different LORAs for various semantic tasks. \textit{Mechanism Difference:} MOE-LORA weights a series of LORA modules fixed in runtime, while our approach directly generates LORA parameters at runtime. This makes our approach fitting the current input image more precisely.

\textbf{Feature Extraction.}
The input image $\mathbf{x}_i$ is fed into the above merged network to produce the feature \( \mathbf{f}_i \in \mathbb{R}^{768 \times h \times w} \), where $h = H/16$ and $w = W/16$ denote the feature resolution.
$\mathbf{f}_i$ is then averaged into the global identity feature for identity comparison.
For multi-slice modalities, such as CT/MRI scans, we extract feature maps from each slice in the scan and further average them as the scan-wise feature. While more advanced inter-slice operations~\cite{pan2022st}\cite{li2022satr} could be employed, we opt for the average operation to maintain the simplicity and efficiency.

\subsection{Learning Rich Medical Priors from MFMs}

\textbf{Motivation.}
With the MedReID loss alone as the learning objective, the model may be biased towards the trivial textures, such as machine noises.
% factors that are not medically relevant, such as machine noise for the same machine, and the patient's usual lying position.
In contrast, medical foundation models (MFMs) pre-trained on massive medical images focus on anatomical characteristics, which is more related to the patient intrinsic identity. This motivates us to transfer the rich medical priors within MFMs to our model.

Considering that local features contain more fine-grained information than global features,
we use the local feature map of MFMs to guide our model.
Furthermore, to close the domain gap between the pre-training task of MFMs and our MedReID task, we propose two strategies, (1) selecting the identity-related key features from the local features, (2) learning the inter-image differential features, instead of the single-image features, as shown in Figure~\ref{fig:framework} (c).

\textbf{Key Feature Selection.}
Given the modality feature $\mathcal{M}_i$ of the image $\mathbf{x}_i$, we use a three-layer MLP to map it into $N$ query tokens $\mathbf{O}_{{i}} = \{\mathbf{O}_{{i}}^1, \mathbf{O}_{{i}}^2, \ldots, \mathbf{O}_{{i}}^N\}$, where $N$ denotes the query number.
The above tokens are modality-specific, enabling flexible handling of key structures in different modalities. For example, key features for Chest X-ray images include ribbon shape, heart size, and clavicle shape, while key features for fundus images include optic disc shape and vessel patterns, etc.

For the $n$th query token $\mathbf{O}_{{i}}^n \in \mathbb{R}^{d}$, we calculate its attention map $\mathbf{A}^n_i$ with the image feature map $\mathbf{f}_i$,
	\vspace{-1mm}
\begin{equation}
	\begin{aligned}
		\small
		\mathbf{A}^n_i &= \text{Softmax}\left(\frac{\mathbf{O}_{{i}}^n  \operatorname{Linear}(\mathbf{f}_i)}{\sqrt{d}}\right) \in \mathbb{R}^{h \times w} ,
	\end{aligned}
\end{equation}
where the feature dimension $d$ is 768, $\operatorname{Linear}$ denotes a linear transformation.
Then, $\mathbf{A}^n_i$ attentively pools the feature map $\mathbf{f}_i$, producing the $n$th key feature $\mathbf{P}^n_i = \sum_{j=o}^{h \times w} \mathbf{A}^n_i[o] \cdot \mathbf{f}_i[o]$, where $o$ denote the spatial position index.
For the features from the MFM, we first choose the MFM from the MFM sets, based on the modality of $\mathbf{x}_i$.
Then, the $n$th key feature is selected in a similar manner, denoted as $\mathbf{Q}^n_i$.

\textbf{Feature Difference Alignment.}
Considering that the MedReID task requires modeling the subtle differences between different images, we propose to align the inter-image feature difference from our model to those of MFM, instead of directly aligning singe-image feature.
Given two medical images, $\mathbf{x}_i$ and $\mathbf{x}_j$, after performing the above feature selection procedures, the $n$th key features from our model are denoted as $\mathbf{P}^n_i$ and $\mathbf{P}^n_j$,
while those from the MFM are denoted as $\mathbf{Q}^n_i$ and $\mathbf{Q}^n_j$.
% As these features are already explicitly aligned , 
Then,
we could use a simple subtraction operation to calculate the $n$th feature differences, which are given by $\mathbf{u}^n = \mathbf{P}^n_i -  \mathbf{P}^n_j$ and $\mathbf{v}^n = \mathbf{Q}^n_i -  \mathbf{Q}^n_j$, respectively, for our model and the MFM, respectively.
Then, we adopt the contrastive loss to align the above features,
	\vspace{-3mm}
\begin{equation}
	\small
	\mathcal{L}_{{med}-{align}}  = \frac{1}{N} \sum_{n=1}^{N}-\log(
	S(\mathbf{u}^n, \mathbf{v}^n)
	),
	\vspace{-3mm}
\end{equation}
where 
\vspace{-1mm}
\begin{equation}
	\small
		S(u^n, v^n)	 = \frac{\exp({\mathbf{u}^n \cdot \mathbf{v}^n}/\tau)}{\exp({\mathbf{u}^n \cdot \mathbf{v}^n}/\tau) + \sum_{k \in \mathcal{N}}^{} \exp({\mathbf{u}^n \cdot \mathbf{v}^k}/\tau)},
  \vspace{-1mm}
\end{equation}
where $\mathcal{N}$ denotes negative samples, which include non-$n$th feature differences of the image pair ($\mathbf{x}_i$,$\mathbf{x}_j$), as well as all feature differences from other image pairs.
$\tau$ denotes the temperature, which is set to 0.07, following MoCo~\cite{chen2021empirical}.

\subsection{Framework Training}
To enable our model to discriminate the medical images from different patients, while also of rich medical priors, we adopt the following loss function:
\vspace{-2mm}
\begin{equation}
	\mathcal{L} = \underbrace{\mathcal{L}_{id-classify} + \mathcal{L}_{tri}}_{\text{Identity terms}} + \underbrace{\lambda \mathcal{L}_{\text{med-align}}}_{\text{Medical term}},
    \vspace{-2mm}
\end{equation}
where $\mathcal{L}_{id-classify}$ is the cross-entropy loss for patient ID classification, $\mathcal{L}_{tri}$ denotes the triplet loss
with soft margin, following~\cite{he2021transreid}.
$\lambda$ denotes the balancing weight.

%\vspace{-2mm}
\section{Experiments}

\subsection{Model Details}

\textbf{Implementation Details.}
During training, we apply random flipping, random cropping, random erasing~\cite{zhong2020random}, and random slice sampling for data augmentation. Specifically, random flipping involves horizontal and vertical flips, while random cropping randomly crops the patches of size 224 $\times$ 224 from the original image.
Random slice sampling denotes randomly selecting 8 slices of the CT scans.
For each training batch, all images belong to the same modality.
$\lambda$ is set to 0.01.
The rank number of the generated parameters is set to 16.
The group number of the last linear layer of FFN-PNet and Att-PNet is set to 64.
At test time, we resize the shorter side of the images to 256 and then center-crop the middle 224$\times$224 region. For multiple-slice scans, we uniformly sample 8 slices.
The initial learning rate is set to 1e-5 and is gradually decayed with the cosine annealing strategy~\cite{loshchilov2016sgdr}. The total number of training steps is 300,000.
The mini-batch size is 196 for single-image medical imaging, while 24 for multiple-slice medical sequences.
We utilize the AdamW optimizer~\cite{loshchilov2017decoupled} implemented in PyTorch~\cite{paszke2019pytorch} with CUDA support. The values of $\beta_1$ and $\beta_2$ are set to 0.9 and 0.999, respectively. The weight decay is set to 0.05.
The entire training process takes about two days on a machine equipped with four NVIDIA RTX 4090 GPUs.

\textbf{Medical Foundation Models.}
For X-ray modality, we adopt the Med-Unic~\cite{wan2024med,wan2024med_github}.
% of which the official pre-trained model weights are provided in~\cite{}.
For CT modality, we adopt the CT-CLIP~\cite{hamamci2024foundation_ctclip,hamamci2024ctclip}.
% of which the official pre-trained model weights are provided in~\cite{hamamci2024ctclip}.
For fundus image modality, we adopt the RetFound~\cite{zhou2023foundation,hamamci2024ctclip}.
% of which the official pre-trained model weights are provided in~\cite{hamamci2024ctclip}.
For histopathology modality, we adopt the CHIEF~\cite{wang2024pathology,wang2024pathology_code}.

\textbf{Evaluation Metrics.}
Following ~\cite{he2021transreid}, we adopt the cumulative matching characteristics (CMC)~\cite{bolle2005relation} at Rank-1 (R1), i.e., CMC-R1, to evaluate the ReID performance.

\subsection{Datasets}
\textbf{Training and Internal Validation Sets.}
We re-organize the public datasets with multiple images per patient, excluding those with less than two images, to ensure each patient has at least one query and target images for re-identification.
The re-organized datasets include,
(1) 111333 \textbf{\color{gray} X-ray} images from \textit{MIMIC-X}~\cite{johnson2019mimic}.
% 106333/5000 \textbf{\color{gray} X-ray} images from 27852/1000 patients of \textit{MIMIC-X}~\cite{johnson2019mimic} for training/validation,
(2) 2460 \textbf{\color{gray} lung CT} scans from \textit{CCII}~\cite{zhang2020clinically}.
% 1473/987 \textbf{\color{gray} lung CT} scans from 601/401 patients of \textit{CCII}~\cite{zhang2020clinically} for training/validation,
(3) 211 \textbf{\color{gray} abdominal CT} scans from \textit{HCC-TACE}~\cite{moawad2021multimodality}.
% 127/84 \textbf{\color{gray} abdominal CT} scans from 63/42 patients of \textit{HCC-TACE}~\cite{moawad2021multimodality} for training/validation,
and 
(4) 35126 \textbf{\color{gray} eye fundus} images \textit{EyePACS}
% 35126 \textbf{\color{gray} eye fundus} images from 17563 patients of \textit{EyePACS}~\cite{diabetic-retinopathy-detection} for training. We do not preserve samples for validation, because there are many other funds datasets for evaluation.
(5) 6068 \textbf{\color{gray} eye fundus} images from \textit{ODIR}~\cite{odir2019}.
% 3640/2428 \textbf{\color{gray} eye fundus} images from 1820/1214 patients of \textit{ODIR}~\cite{odir2019} are for training/validation. Compared to EyePACS which mostly consists of diabetic retinopathy diseases, ODIR includes more other eye diseases such as Glaucoma and Cataracts.
(6) 542 \textbf{\color{gray} histopathology} images from \textit{LUAD}~\cite{national9clinical}.
% 267/275 \textbf{\color{gray} histopathology} images from 62/63 patients of \textit{LUAD}~\cite{national9clinical} are for training/validation.
The train/validation splitting protocols and dataset details are provided in the supplementary material.

\textbf{External Validation Sets.}
We also evaluate our model on six external validation sets, the results of which can reflect the model's generalization capability.
(1) To build external \textbf{\color{gray} X-ray} set, we sample 6569 images of 1000 patients from \textit{Chest-X}~\cite{wang2017chestx}.
(2) To build \textbf{\color{gray} abdominal CT} set, we sample 239 CT scans of 70 patients from \textit{KIRC}~\cite{akin2016cancer}.
(3) As another \textbf{\color{gray} abdominal CT} set, we sample 194 CT scans of 56 patients from \textit{LIHC}~\cite{akin2016cancer}. It is worth mentioning that a little proportional of LIHC contains the MRI images.
(4) To build \textbf{\color{gray} brain MRI} set, we use all 55 MRI scans of 20 patients from \textit{OASIS2}~\cite{marcus2010open}.
(5) To build \textbf{\color{gray} eye fundus} image set, we use 700 fundus images of 350 patients from \textit{Mess2}~\cite{decenciere2014feedback}.
(6) As another \textbf{\color{gray} eye fundus} image set, we use all 521 images of 144 patients from \textit{GRAPE}~\cite{huang2023grape}.

\begin{table*}[htbp]
	\centering
	\tabcolsep=1.2mm
	\renewcommand{\arraystretch}{0.1}
	\resizebox{0.99\textwidth}{!}{%
		\begin{tabular}{lcccccccccccH}
			\toprule
			\multirow{3}{*}{Method} & \multicolumn{11}{c}{Dataset} \\
			\cmidrule{2-12}
			& \multicolumn{1}{c}{\small MIMIC-X} & \multicolumn{1}{c}{\small Chest-X} & \multicolumn{1}{c}{\small CCII} & \multicolumn{1}{c}{ \small HCC-TACE} & \multicolumn{1}{c}{\small KIRC} & \multicolumn{1}{c}{\small LIHC} & \multicolumn{1}{c}{\small OASIS2} & \multicolumn{1}{c}{\small Mess2} & \multicolumn{1}{c}{\small ODIR} & \multicolumn{1}{c}{\small GRAPE} & \multicolumn{1}{c}{\small LUAD} &  ISIC20 \\
			& \multicolumn{1}{c}{\small X-Ray} & \multicolumn{1}{c}{\small X-Ray} & \multicolumn{1}{c}{\small Lung-CT} & \multicolumn{1}{c}{\small Ab-CT} & \multicolumn{1}{c}{\small Ab-CT} & \multicolumn{1}{c}{\small Ab-CT} & \multicolumn{1}{c}{\small Br-MRI} & \multicolumn{1}{c}{\small Fundus} & \multicolumn{1}{c}{\small Fundus} & \multicolumn{1}{c}{\small Fundus} & \multicolumn{1}{c}{\small Histo} &  Skin \\
			\midrule
			\multicolumn{13}{l}{\textcolor{gray} {\small \textit{\textbf{Visual Foundation Models}}}} \\
			ImageNet-Sup~\cite{dosovitskiy2020image} & 34.10 & 39.90 & 84.04 & 50.00 & 47.14 & 26.78 & 47.99 & 47.14 & 32.70 & 44.30 & 29.13 & 13.25 \\
			MoCoV3~\cite{chen2021empirical} & 45.10 & 46.50 & 84.79 & 45.24 & 46.43 & 30.36 & 50.00 & 56.86 & 42.26 & 59.93 & 47.24 & 12.75 \\
			DINOv2~\cite{oquabdinov2} & 36.40 & 37.60 & 91.52 & 50.00 & 42.86 & 28.57 & 46.00 & 36.00 & 23.72 & 41.37 & 42.52 & 8.50 \\
			BEITv2~\cite{peng2022beit} & 35.10 & 35.30 & 89.53 & 52.38 & 25.00 & 30.36 & 70.00 & 52.00 & 37.89 & 54.40 & 45.67 & 12.25 \\
			CAE~\cite{chen2024context}  & 36.20 & 32.40 & 71.32 & 45.24 & 28.57 & 21.43 & 40.00 & 41.43 & 28.34 & 50.16 & 47.24 & 15.50 \\
			MAE~\cite{he2022masked} & 23.80 & 23.10 & 68.33 & 35.71 & 32.14 & 23.21 & 15.99 & 47.14 & 23.15 & 33.55 & 30.71 & 10.25 \\
			MaskFeat~\cite{wei2022masked} & 9.20 & 11.60 & 19.95 & 28.57 & 17.86 & 14.29 & 10.00 & 20.00 & 8.32 & 16.94 & 14.17 & 3.50 \\
			MoCoV3$^\dag$ &84.20 & 64.00 & 92.52 & 71.43 & 46.43 & 33.93 & 56.00 & 70.99 & 65.90 & 67.43 & 51.97 & 23.00 \\
			MAE$^\dag$ &88.20 & 68.60 & 93.27 & 76.19 & 57.14 & 41.07 & 60.00 & 72.57 & 61.12 & 60.91 & 45.67 & 32.25 \\
			\midrule
			\multicolumn{13}{l}{\textcolor{gray} {\small \textit{\textbf{Visual-Language Foundation Models}}}} \\
			Align~\cite{jia2021scaling} & 0.40 & 0.90 & 43.39 & 4.76 & 17.86 & 12.50 & 0.00 & 13.71 & 3.38 & 7.82 & 5.51 & 2.50 \\
			BLIP~\cite{li2022blip}  & 3.10 & 4.80 & 79.05 & 21.42 & 25.00 & 14.29 & 10.00 & 33.43 & 10.54 & 17.59 & 24.41 & 6.25 \\
			CLIP~\cite{radford2021learning} & 33.10 & 31.60 & 93.02 & 45.24 & 35.71 & 28.57 & 68.00 & 41.14 & 30.15 & 50.81 & 46.46 & 10.00 \\
			CLIP$^\dag$ &92.30 & 73.00 & 93.52 & 69.05 & \cellcolor{lightgray}  57.14 & \cellcolor{lightgray}  51.79 & \cellcolor{lightgray}  68.00 & 73.71 & 66.06 & 60.52 & 40.94 &  33.17 \\
			\midrule
			
			\multicolumn{13}{l}{\textcolor{gray} {\small \textit{\textbf{Object ReID Model}}}} \\
			TransReID~\cite{he2021transreid} & 29.30 & 33.90 & 88.78 & 33.33 & 39.29 & 26.79 & 69.99 & 42.29 & 30.89 & 36.81 & 30.71 & 11.50 \\
						TransReID$^\dag$ & 86.80 & 68.60 & 93.52 & \cellcolor{lightgray}  80.95 & 47.14 & 39.29 & 64.00 & 74.00 & 65.52 & 60.36 &   54.33 & 29.25 \\
			\midrule
			\multicolumn{13}{l}{\textcolor{gray} {\small \textit{\textbf{Medical Foundation Models}}}} \\
			BioMedClip~\cite{zhang2023biomedclip}  & 25.20 & 24.00 & 82.04 & 40.48 & 32.14 & 26.79 & 32.00 & 23.14 & 19.44 & 27.68 & 33.07 & 8.00 \\
			RetFound~\cite{zhou2023foundation} & 12.10 & 15.00 & 61.85 & 35.71 & 39.29 & 16.07 & 15.99 & 53.71 & 28.83 & 35.50 & 25.98 & 10.25 \\
			CT-CLIP~\cite{hamamci2024foundation_ctclip} & 3.80 & 5.30 & 87.03 & 9.52 & 33.14 & 13.51 & 5.99 & 33.14 & 17.79 & 16.61 & 16.54 & 1.99 \\
			Med-Unic~\cite{wan2024med} & 48.70 & 44.90 & 77.06 & 33.33 & 32.14 & 25.00 & 23.99 & 27.71 & 21.75 & 35.83 & 28.35 & 10.25 \\
			BioMedClip$^\dag$  & 20.10 & 19.00 & 83.04 & 52.38 & 25.00 & 26.79 & 36.00 & 28.57 & 18.62 & 27.69 & 42.52 & 10.99 \\
			RetFound$^\dag$  &54.80 & 42.80 & 92.27 & 66.67 & 28.57 & 35.71 & 50.00 & \cellcolor{lightgray} 74.14 & \cellcolor{lightgray}  66.70 & \cellcolor{lightgray}  61.10 & 37.80 & 25.75 \\
			CT-CLIP$^\dag$ & 19.70 & 19.70 & \cellcolor{lightgray}  94.04 & 47.62 & 21.43 & 28.57 & 37.99 & 29.71 & 19.93 & 29.97 & 42.52 & 10.00 \\
			Med-Unic$^\dag$ & \cellcolor{lightgray} 92.90 &  74.30 & 69.08 & 57.14 & 39.29 & 35.71 & 41.99 & 24.57 & 16.39 & 22.15 & 25.20 & 5.25 \\
			\midrule

			\multicolumn{13}{l}{\textcolor{gray} {\small \textit{\textbf{Modality-specialized MedReID Models}}}} \\
			Packh{\"a}user~\etal~\cite{packhauser2022deep}& 92.42 & \cellcolor{lightgray} 88.21 & 68.63 & 45.24 & 35.7 & 32.11 & 36.02 & 23.74 & 15.18 & 23.77 & 29.19 & 3.75  \\
			Ganz~\etal~\cite{ganz2025re} & 11.40 & 11.90 & 53.62 & 33.33 & 39.29 & 33.93 & 28.00 & 27.43 & 22.65 & 25.08 & \cellcolor{lightgray} 56.76 & 7.00 \\
			\midrule
			\multicolumn{13}{l}{\textcolor{gray} {\small \textit{\textbf{All-in-One MedReID Models}}}} \\
			\rowcolor{gray!25}
			Ours & \textbf{96.89} & \textbf{91.49} & \textbf{95.01} & \textbf{88.09} & \textbf{82.68} & \textbf{76.82} & \textbf{85.00} & \textbf{85.71} & \textbf{71.34} & \textbf{71.00} &\textbf{68.75}& \textbf{30.75} \\
			\bottomrule
		\end{tabular}%
	}
	\vspace{-1mm}
 \caption{
		Comparison of different approaches on medical image re-identification in terms of CMC-R1.
		$^\dag$ indicates the model is further tuned on the medical datasets same as ours.
		%		 for a fair comparison.
		MIMIC and ChestX indicate the MIMIC-CXR and ChestX-Ray14 datasets. `Ab-' and `Br-' denotes the `Abdominal' and `Brian'.
		All models adopt the ViT-Base~\cite{dosovitskiy2020image} architecture with a similar parameter number, for a fair comparison.
		The best and the second best results are marked with \colorbox{lightgray}{\textbf{gray bold}} and \colorbox{lightgray}{{gray}}, respectively.
	}
	\label{tab:sota_reid}
 
	\vspace{-5mm}
\end{table*}

\subsection{Results}

\textbf{MedReID Benchmark.}
As shown in Table~\ref{tab:sota_reid}, we evaluate various visual foundation models,
visual-language foundation models,
Person-ReID model,
medical foundation models,
and single-modality MedReID models.
To fully release their potential, we fine-tune some representative models using our training datasets, ensuring a fair comparison.

For \textit{visual foundation models}, contrastive learning approaches like MoCoV3 and DINOv2 achieve decent performance, with accuracies of 84.79\% and 91.52\% on the CCII (Lung-CT) dataset, respectively. In contrast, masked learning models such as MAE and MaskFeat perform much worse, achieving only 68.33\% and 19.95\% on the same dataset. However, after fine-tuning for the MedReID task, MAE$^\dag$ outperforms MoCoV3$^\dag$ on most datasets. These findings align with previous research \cite{oquabdinov2,he2022masked}, i.e., contrastive features are more linearly separable when being directly deployed, while MAE-style models excel after adaptation due to their more powerful representations.

For \textit{visual-language foundation models}, CLIP consistently outperforms other methods by a substantial margin, achieving 93.02\% accuracy on CCII and 70.00\% on OASIS2. In contrast, Align and BLIP perform much worse, with accuracies below 20\% on OASIS2. These results highlight that CLIP, trained on approximately one billion image-text pairs, learns highly generalizable visual features. After further tuning, the fine-tuned CLIP$^\dag$ shows another significant improvement, surpassing both MoCoV3$^\dag$ and MAE$^\dag$ models by a large margin.
For example, on the Chest-X dataset, CLIP$^\dag$, MoCoV3$^\dag$, and MAE$^\dag$ achieve, 73.00\%, 64.00\%, and 68.60\%, respectively.

\textit{Person ReID} method TransReID has generally produced suboptimal results when applied to medical images, largely attributed to the substantial domain gap between person images and medical images. After fine-tuning, TransReID$^\dag$ improves somewhat, but still lags far behind CLIP$^\dag$.

For \textit{medical foundation models}, BioMedClip performs much inferior to CLIP, due to the smaller training dataset PMC-15M. Specialized models like Med-UniC achieve decent performance in their training modality, such as 48.70\% accuracy on X-ray images, but perform poorly on other modalities like fundus and CT. This is similar to CT-CLIP and RetFound.
After fine-tuning, CT-CLIP$^\dag$, RetFound$^\dag$, and Med-UniC$^\dag$ show a further performance boost on the modalities consistent with their pre-training dataset, demonstrating that their pre-trained medical priors are beneficial for the ReID task, but perform unsatisfactorily on other modalities. For example, RetFound$^\dag$ achieves 74.14\% on Mess2 (fundus), outperforming the strong CLIP$^\dag$, but only 42.80\% on Chest-X (X-ray).

\textit{Single-modality MedReID} methods \cite{packhauser2022deep,ganz2025re} fail to generalize to the modalities out of the training scope. For instance, the X-ray ReID model \cite{packhauser2022deep} attains 92.42\% accuracy on MIMIC-X (X-ray) but only 15.18\% on ODIR (fundus).
In contrast, we outperform them by a large margin, due to learning and combining identity cues from several diverse-modality training sources. Additionally, we surpass fine-tuned medical foundation models, such as RetFound$^\dag$ and Med-UniC$^\dag$, by inheriting and adapting their medical priors to the MedReID problem.
Our approach also surpasses various visual foundation models, achieving state-of-the-art performance across all modalities and datasets.

We benchmark eight \textit{large visual-language} model on the medical ReID task. The results are detailed in the supplementary material. Our approach also demonstrates obvious superiority, achieving 98.80\% accuracy on Chest-X, while QWen-VL-Max and GPT-4o only achieves 76.80\% and 62.50\%, respectively.

Finally, we study the cross-modality capability of our model.
We evaluate models on a licensed private dataset of 1814 respiratory patients with paired Chest X-ray and CT images. Our all-in-one model learns to associate patient ID across modalities, achieving 87.28\% accuracy (Tab. R1), outperforming single-modality-only models. This suggests the all-in-one paradigm benefits cross-modality ReID. The fine-tuning further improves the result to 94.38\%.

\begin{table}[thbp]
	\centering
	\tabcolsep=1.2mm
	\renewcommand{\arraystretch}{0.1}
	\resizebox{0.99 \linewidth}{!}{%
	\begin{tabular}{|c|c|c|c|}
		\hline
		MaMI(X-ray only) & MaMI(CT only) & MaMI(Ours) & MaMI(Tuned) \\
		\hline
		76.42\% & 78.19\% & 87.28\% & 94.38\% \\
		\hline
	\end{tabular}%
}\vspace{-2mm}
	\caption{Cross-modality ReID: Using X-ray images to retrieve matching CT images of the same patient, on the test set of the respiratory dataset. MaMI (Ours) refers to our all-in-one model trained without cross-modality image pairs. MaMI (Tuned) denotes fine-tuning on the dataset's cross-modality image pair.}
	\label{tab:cross-modal-reid}
\end{table}

\textbf{Application I: Longitudinal Personalized Healthcare.}
Further, we consider a realistic scenario where patients' past medical images are not under good management. Given the current image, we use MaMI to retrieve relevant historical images and combine them with the current image for diagnosis. Notably, only the images themselves are utilized, without any historical labels. To integrate features from multiple historical images, we employ a simple MLP.

\begin{figure}[!t]
		% \vspace{-2mm}
	\centering
	\tabcolsep=0.1mm
	\begin{tabular}{cc}
			\includegraphics[width=0.48 \linewidth]{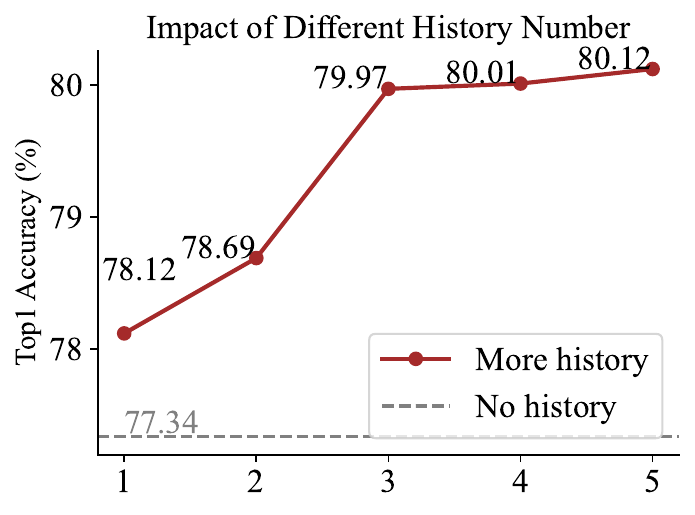}
		 &
			\includegraphics[width=0.48 \linewidth]{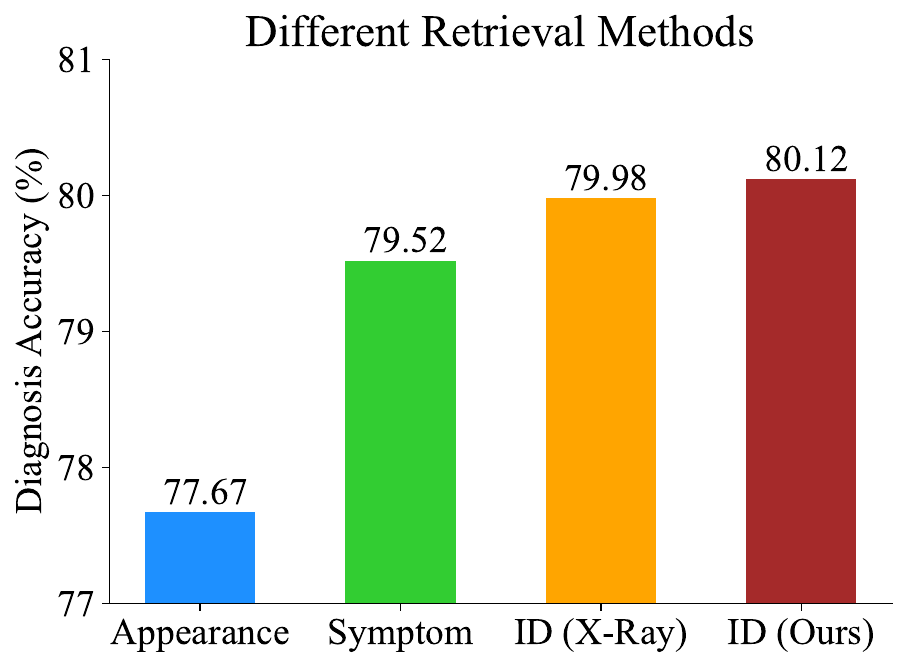}
	\end{tabular}
	\vspace{-3mm}
	\caption{
	Impact of the historical image number on diagnosis outcome. 
	We use the proposed MaMI to collect the historical image, as the auxiliary information, to aid the diagnosis.
	}
		\vspace{-6mm}
	\label{fig:long_outcome}
\end{figure}

As shown in Figure~\ref{fig:long_outcome} \textit{left}, the enhancement through historical image retrieval boosts diagnostic accuracy, due to more longitudinal observations. Specifically, when retrieving five historical images, the accuracy increases from 77.34\% to 80.12\%, a gain of 2.78\%. This demonstrates that MaMI can effectively enhance clinical utility by retrieving relevant historical data from unstructured archives.
We further compare different image retrieve approaches, as shown in Figure~\ref{fig:long_outcome} \textit{right}. Our approach consistently outperforms the appearance(DINOv2~\cite{oquabdinov2})-based, symptom(Med-Unic~\cite{wan2024med})-based, and X-ray-specialized ReID (Packh{\"a}user~\etal~\cite{packhauser2022deep}) methods.

\begin{table}[!thbp]
\vspace{-2mm}
	\centering
	\tabcolsep=1.2mm
	\renewcommand{\arraystretch}{0.6}
	\resizebox{0.99\columnwidth}{!}{%
		\begin{tabular}{l||c|c|c|c|c}
			\hline
			& MaMI&MAE$^\dag$ & CLIP$^\dag$ & Med-Unic$^\dag$& \cite{packhauser2022deep}\\
			\hline
			Original&91.49\% & 68.60\% & 73.00\%&  74.30\%  & 88.21\% \\
   	\rowcolor{gray!25}
			Protected&21.23\% & 14.52\% & 11.86\%&  13.94\%  & 15.68\% \\
			\hline
		\end{tabular}%
	}
	\vspace{-2mm}
 \caption{
 MedReID on the protected Chest-X dataset.
 The medical visual cues are detected by our MaMI model, while the privacy removal images can resist attacks from other ReID models.
 $^\dag$ indicates the model is further tuned on the medical datasets same as ours, for a fair comparison.
 }
	\label{tab:protect_id}
 % \vspace{-3mm}
\end{table}

\textbf{Application II: Privacy Protection.}
We adopt a simple U-Net~\cite{ronneberger2015u} to predict identity-related visual cues and remove them from the original images. The training objective is to minimize the identity similarity distance between the identity-removed image and the original image, while maximizing their medical feature similarity. Details are in the supplementary material.
The identity-removal U-Net is trained on the MIMIC-X dataset and evaluated on the Chest-X dataset. As shown in Table~\ref{tab:protect_id}, the protected images resist re-identification attacks from various ReID models.
We also train disease classification models on both the original and privacy-protected datasets. The accuracies are 81.24\% and 80.67\%, respectively, indicating that the privacy-protected images preserve the data utility well.

% \vspace{-2mm}
\subsection{Model Analysis}\label{sec:method_ana}

\textbf{Framework-level Ablation Study.}
As shown in Table~\ref{tab:ablation_study_framework}, the baseline model M$_{\text{base}}$, which naively fine-tunes the CLIP model on our multi-modality training dataset, results in the poorest performance.
Introducing the modality-adaptive component ComPA, the resulting M$_{\text{compa}}$ achieves substantial gains on various modalities, i.e., 4.31\% and 11.91\% gains on MIMIC-X (X-ray) and HCC-TACE (CT),
due to handling inter-modality heterogeneity.

\begin{table}[!t]
	\centering
	\tabcolsep=0.5mm
%	\vspace{-3mm}
		\renewcommand{\arraystretch}{0.9}
	\resizebox{0.99\columnwidth}{!}{%
	\begin{tabular}{l||cc|cc|cc}
		\hline
		\multirow{2}{*}{Model} &\multirow{2}{*}{ComPA} & \multirow{2}{*}{MFMs} & \multicolumn{2}{c|}{Internal Validation} & \multicolumn{2}{c}{\textit{External Validation}}\\
		\cline{4-7}
		& & & MIMIC-X &HCC-TACE & Chest-X  & GRAPE \\
		\hline
		M$_{\text{base}}$ &\xmark & \xmark & 92.29  & 69.04& 86.21 & 66.51 \\
		M$_{\text{compa}}$ &\checkmark & \xmark & 96.60  & 80.95& 89.35 & 69.45 \\
        \hline
			\rowcolor{gray!25}
		M$_{\text{ours}}$ &\checkmark & \checkmark &\textbf{ 96.89} & \textbf{88.09} & \textbf{91.49} & \textbf{71.00} \\
		\hline
	\end{tabular}%
}
	\vspace{-2mm}
\caption{Ablation study on the two core designs, including the continuous-modality parameter adapter (ComPA) and the medical prior learning from a bag of medical foundation models (MFMs).
}
	\label{tab:ablation_study_framework}
 	\vspace{-4mm}
\end{table}

Further alignment with Medical Foundation Models (MFMs) to enrich the model's medical prior, resulting in M$_{\text{ours}}$, yields additional performance gains, especially in data-scarce situations. On the HCC-TACE dataset, which contains only 127 training samples, performance increases from 80.95\% to 88.09\%. This demonstrates that MFMs mitigate the data scarcity issue common in medical imaging.
On external datasets (Chest-X and GRAVE), M$_{\text{ours}}$ surpasses M$_{\text{compa}}$ by 2.14\% and 1.55\%, respectively. The good results on external validation datasets highlight the generalizability of features derived from MFMs.

In summary, both ComPA and MFM alignment are crucial. The ComPA improves overall performance on various modalities, while MFM alignment mitigates the data-scarcity problem and enhances generalization capability.

\textbf{Ablation Study on ComPA.}
We further investigate if all designs within ComPA are necessary.
As shown in Table~\ref{tab:ablation_study_compa}, without considering any modality specificity, the baseline model M$_{\text{mod-no}}$ achieves 92.36\% and 86.42\% on MIMIC-X and Chest-X datasets, respectively.

\begin{table}[!t]
%\vspace{-3mm}
\small
	\centering
	\tabcolsep=1.0mm
		\renewcommand{\arraystretch}{0.6}
	
	\resizebox{0.99\columnwidth}{!}{%
		\begin{tabular}{l||ccc|c|c}
			\hline
			Model & Adaptive&Modality &Codebook& MIMIC-X &Chest-X \\
			\hline
			M$_{\text{mod-no}}$ & \xmark &-&-& 92.36\%&86.42\% \\
			M$_{\text{mod1}}$ & \xmark &Discrete &-& 94.67\% & 87.87\% \\
			M$_{\text{mod2}}$ & \checkmark &Continuous&\xmark & 96.78\% & 90.12\% \\
   \hline
   	\rowcolor{gray!25}
			M$_{\text{ours}}$ & \checkmark &Continuous&\checkmark & \textbf{96.89\%} & \textbf{91.49\%} \\
			\hline
		\end{tabular}%
	}
	\vspace{-2mm}
 \caption{Ablation study on ComPA.
All models are incorporated with the $\mathcal{L}_{med-align}$ loss item for rigorous ablation.
	}
	% \vspace{-3mm}    
	\label{tab:ablation_study_compa}
\end{table}

With the discrete modality labels, such as X-ray as 0, Lung CT as 1, Abdominal CT as 2, etc, as the input condition, the produced M$_{\text{mod1}}$ substantially improves upon M$_{\text{mod-no}}$ by 1.45\% on Chest-X, proving that modality information is critical for a unified MedReID model.
Further introducing instance-adaptive continuous modality design, M$_{\text{mod2}}$ surpasses M$_{\text{mod1}}$ by another 2.11\% on MIMIC-X, indicating that the continuous design better captures data nuances.
The introduction of codebook design leads to further improvements, particularly on the external validation set Chest-X (+1.37\%). This suggests that the codebook enhances the model's out-of-domain generalizability.

\begin{figure}[!t]
 % \vspace{-2mm}
	\centering
	\tabcolsep=-0.2mm
	\small
	\begin{tabular}{cc}
	\includegraphics[width=0.50 \linewidth]{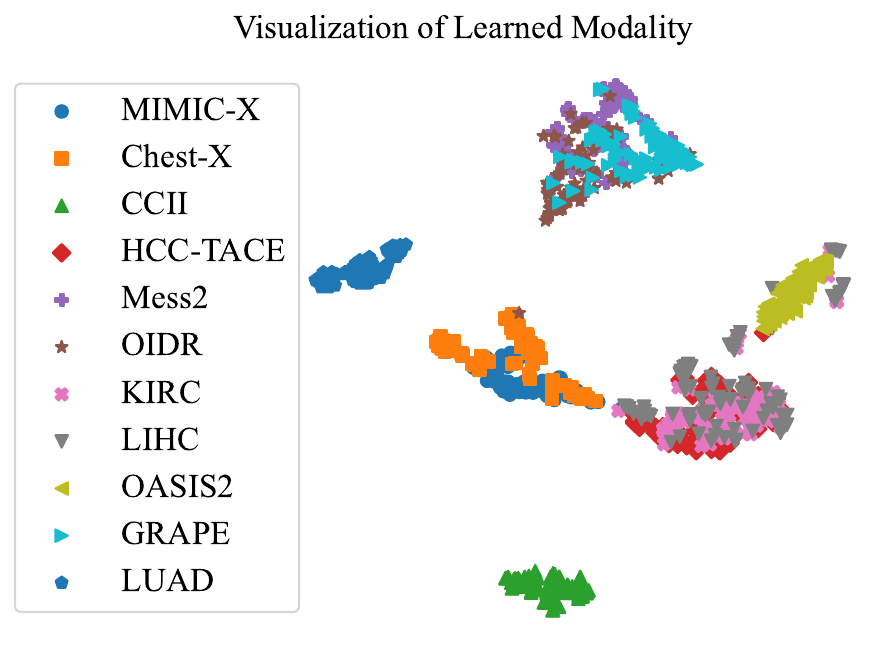}&	
  \includegraphics[width=0.45 \linewidth]{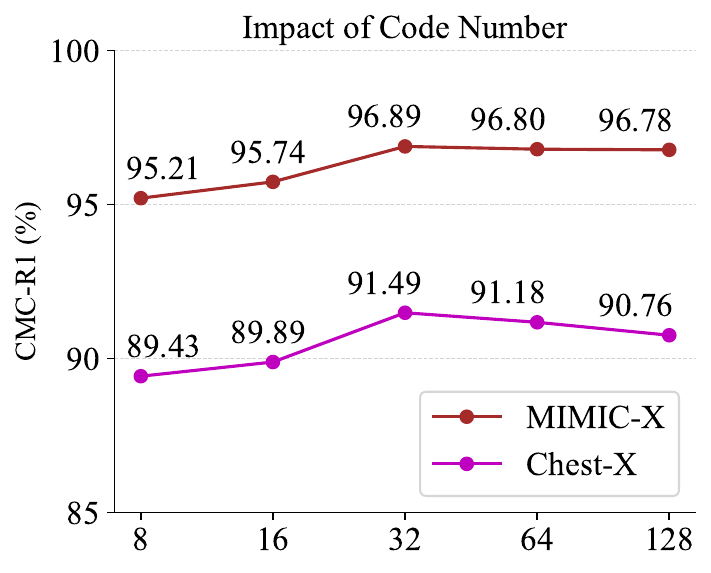}\\
  % \includegraphics[width=0.32 \linewidth]{code/group_number.pdf}\\
  
  % (a)&(b)
	\end{tabular}
 \vspace{-2mm}
	\caption{
\textit{Left:} t-SNE map of the learned continuous modality.
\textit{Right:} Impact of the code number of the codebook within ComPA.
	}
	\label{fig:tsne_continous_modality}
 \vspace{-2mm}
\end{figure}

\begin{table}[!t]
% \vspace{-2mm}
\small
\tabcolsep=0.5mm
    \centering
    \begin{minipage}[t]{0.49\columnwidth}
        \centering
         \resizebox{0.95\columnwidth}{!}{%
        \begin{tabular}{cccc}
            \hline
            Rank & 8&16&32 \\
            \hline
            Chest-X  & 85.56\% & \textbf{91.49\%} &91.50\% \\
            \hline
        \end{tabular}%
        }
        % \subcaption{}
    \end{minipage}%
    \hfill % 在两个 minipage 之间添加水平填充
    \begin{minipage}[t]{0.49\columnwidth}
        \centering
        \resizebox{0.95\columnwidth}{!}{%
        \begin{tabular}{cccc}
            \hline
            Group & 32&64&128 \\
            \hline
            Chest-X  & 91.45\% & \textbf{91.49\%} &90.52\% \\
            \hline
        \end{tabular}%
        }
        % \subcaption{}
    \end{minipage}%
    \hfill % 在两个 minipage 之间添加水平填充
    \vspace{-2mm}
    \caption{Impact of (\textit{left}) rank number of the generated parameters, and (\textit{right}) group number of the parameter-generation layers.}
      \vspace{-5mm}
    \label{fig:hyper_param_compa}
\end{table}

Next, we visualize the learned instance-adaptive modality features by t-SNE\cite{van2008visualizing}. Figure \ref{fig:tsne_continous_modality} \textit{left} shows a clear separation between different modalities (MIMIC-X and Mess2), while the same modality datasets (Mess2 and OIDR) cluster closely.
We observe that LIHC contains some outliers, as a small proportion of LIHC cases are abdominal MRI scans instead of CT scans.
Notably, our model autonomously groups OASIS2 MRI images, despite not training with the brain-MRI data, underscoring the high robustness of our modality representation.
Then, we train different variant models by tuning the codebook size. As shown in Figure~\ref{fig:tsne_continous_modality} \textit{right}, a small code size such as 8 severely reduces performance on all datasets (96.89\% $\rightarrow$ 95.21\% for MIMIC-X and 91.49\% $\rightarrow$ 89.43\% for Chest-X), while a large codebook size such as 128 mainly degrades the model generalizability, i.e., 91.49\% $\rightarrow$ 90.76\% on external Chest-X.

We further investigate the impact of other ComPA hyper-parameters. As shown in Table~\ref{fig:hyper_param_compa}, a small rank constrains model representation, while ranks larger than 16 lead to performance saturation and increased computational cost. For group number, performance is stable at 32 and 64 but degrades at 128 due to much-reduced parameters of FFN- and Att-PNet.
For $\lambda$, our model achieves very similar performance for 0.1 and 0.01 (91.32\% \textit{v.s.} 91.49\% on Chest-X), but inferior performance 89.82\% for 0.001, due to the too loose medical prior regularization.

\textbf{Learning Strategy of Medical Priors.}
As shown in Table~\ref{tab:ablation_medical_prior}, compared to the baseline model M$_{\text{med-no}}$ (no medical priors), introducing global medical priors (M$_{\text{med1}}$) yields minimal gains, as global features fail to capture subtle identity information. Naive local priors (M$_{\text{med2}}$) marginally surpass M$_{\text{med1}}$ by 0.23\%. After the modality-specific feature selection operation, feature semantics is significantly improved, reflected by a substantial gain of 0.92\%.

Replacing single-image feature alignment with inter-image feature relation alignment, where the relation feature is obtained by concatenating the features from different images and feeding them into a three-layer MLP, further boosts performance by 1.20\% in M$_{\text{med4}}$. Finally, substituting the MLP with a subtraction operation in M$_{\text{ours}}$ enforces the model's focus on subtle image differences, achieving a final performance of 91.49\%.
This proves that modeling the inter-image relationship is crucial for the ReID problem, regardless of the specific relationship operator. Both the MLP and our subtraction operation achieve good results.

\begin{table}[!t]
% \vspace{-2mm}
	\centering
	\tabcolsep=1.1mm
		\renewcommand{\arraystretch}{0.6}
	
	\resizebox{0.99\columnwidth}{!}{%
		\begin{tabular}{l||ccc|c}
    \hline
    Model & Feature &Inter-Image&Relation Operator& Chest-X \\
    \hline
    M$_{\text{med-no}}$&-&-&-& 88.54\% \\
    M$_{\text{med1}}$&Global&-&-&88.87\% \\
    M$_{\text{med2}}$ &Local&\xmark&-& 89.10\% \\
    M$_{\text{med3}}$ &Selected&\xmark&-& 90.02\% \\
     M$_{\text{med4}}$ &Selected&\checkmark&MLP& 91.22\% \\
     \hline
     	\rowcolor{gray!25}
    M$_{\text{ours}}$  &Selected&\checkmark&Subtraction& \textbf{91.49\%}\\
    \hline
		\end{tabular}%
	}
	\vspace{-2mm}
 \caption{Strategies of learning medical priors from MFMs.
 }
	\vspace{-5mm}
	\label{tab:ablation_medical_prior}
\end{table}

\textbf{Model Complexity.}
With batch size 128, the inference time of our model is 151.56 ms on a machine with an NVIDIA 4090 GPU, compared to 141.21 ms for the vanilla ViT-Base. The ComPA module only additionally consumes 10ms, as it primarily consists of several simple MLPs to compute modality-specific parameters. Given its brought substantial result gains, this minor increase in latency is justified. Further, the MFM alignment procedure incurs no inference cost, as it only regularizes the training procedure.

\vspace{-2mm}
\section{Conclusion}
\vspace{-2mm}
In this paper, we have introduced a comprehensive benchmark and a unified model for a novel MedReID problem, covering a wide range of medical modalities. We have proposed a modality-adaptive architecture to enable a single model to handle diverse medical modalities at runtime. Additionally, we integrate medical priors into our model by exploiting the pre-trained medical foundation models. Our model substantially outperforms all previous approaches.

\noindent \textbf{Acknowledgment} This work was supported by Shanghai Artificial Intelligence Laboratory, National Natural Science Foundation of China (Grant No.72293585, No.72293580, No.62225112), the Fundamental Research Funds for the Central Universities, National Key R\&D Program of China 2021YFE0206700, Shanghai Municipal Science and Technology Major Project (2021SHZDZX0102), and STCSM 22DZ2229005.
 
{
    \small
    \bibliographystyle{ieeenat_fullname}
    \bibliography{main}
}


\end{document}